%% file: main.tex
\documentclass[runningheads]{llncs}
\usepackage[T1]{fontenc}
\usepackage{graphicx}
\usepackage{amsmath}
\usepackage{amssymb}
\usepackage{multirow}
\usepackage{hyperref}
\begin{document}
\title{Language-Based Digital Twins for Elderly Cognitive Assistance}
%
%
\author{Mohammad~Mehdi~Hosseini\inst{1}\orcidID{0009-0000-5387-6039} \and
Mohammad~H.~Mahoor\inst{1}\thanks{Corresponding author: Mohammad H. Mahoor (\email{mohammad.mahoor@du.edu}).}\orcidID{0000-0001-8923-4660} \and Hiroko~H.~Dodge\inst{2}\orcidID{0000-0001-7290-8307}}
\authorrunning{M. Hosseini et al.}
%
\institute{Ritchie School of Engineering and Computer Science, University of Denver, Denver, CO 80208, USA \\
\email{mohammadmehdi.hosseini@du.edu}\\
\email{mohammad.mahoor@du.edu}\\
\and
Department of Neurology, Massachusetts General Hospital, Harvard Medical School, Boston, MA 02114, USA\\
\email{hdodge@mgh.harvard.edu}
}

\maketitle              
\begin{abstract}
Digital twins have emerged as a promising paradigm for personalized healthcare, enabling modeling of individual behavior and health trajectories. In cognitive health, early detection of Mild Cognitive Impairment (MCI) remains challenging, where language and conversational patterns serve as non-invasive biomarkers. In this work, we propose a language-based digital twin framework that leverages large language models (LLMs) to mimic the conversational behavior of elderly individuals by incorporating stylometric cues and contextual metadata.

To evaluate fidelity and cognitive consistency, we introduce a multi-head conditional variational autoencoder (cVAE) that jointly measures reconstruction quality and predicts cognitive scores. Experiments on the I-CONECT dataset show that the digital twin preserves identity-specific characteristics and achieves reconstruction and MoCA prediction errors comparable to real data, while outperforming baseline GPT-generated responses. These results highlight the potential of language-based digital twins as a scalable and non-invasive approach for personalized and continuous cognitive health monitoring.

\end{abstract}
\section{Introduction}
Digital twins are virtual representations of physical entities that are continuously updated using real-world data, enabling simulation, analysis, and prediction of system behavior. Originally developed in engineering, they have emerged as a promising paradigm in healthcare for building personalized and dynamic computational models of individuals by integrating multimodal data such as clinical records, physiological signals, and behavioral observations. The increasing availability of conversational data has further enabled modeling of human cognition through language and interaction patterns. Unlike conventional machine learning models that focus solely on prediction, digital twins support continuous and individualized behavioral modeling, making them particularly suitable for capturing subtle and longitudinal cognitive changes.

Mild Cognitive Impairment (MCI) is a transitional stage between normal aging and dementia, where early detection is critical for timely intervention~\cite{petersen2014mild}. However, traditional diagnostic approaches rely on structured assessments and neuroimaging, which are costly, infrequent, and limited in capturing gradual changes in everyday behavior~\cite{jack2018nia}. Language and speech have emerged as scalable and non-invasive biomarkers of cognitive decline, with features such as lexical diversity, fluency, and pauses correlating with cognitive status~\cite{martinez2021ten}. Naturalistic conversational data, in particular, provide richer and more temporally informative signals than constrained clinical tasks~\cite{lima2025evaluating}.

Despite these advances, most existing approaches focus on prediction rather than personalized and holistic modeling. Similarly, prior digital twin research in healthcare primarily targets physiological or population-level representations, with limited attention to individual-specific linguistic and behavioral characteristics.

\begin{figure}[t]
\centering
\includegraphics[width=0.9\textwidth]{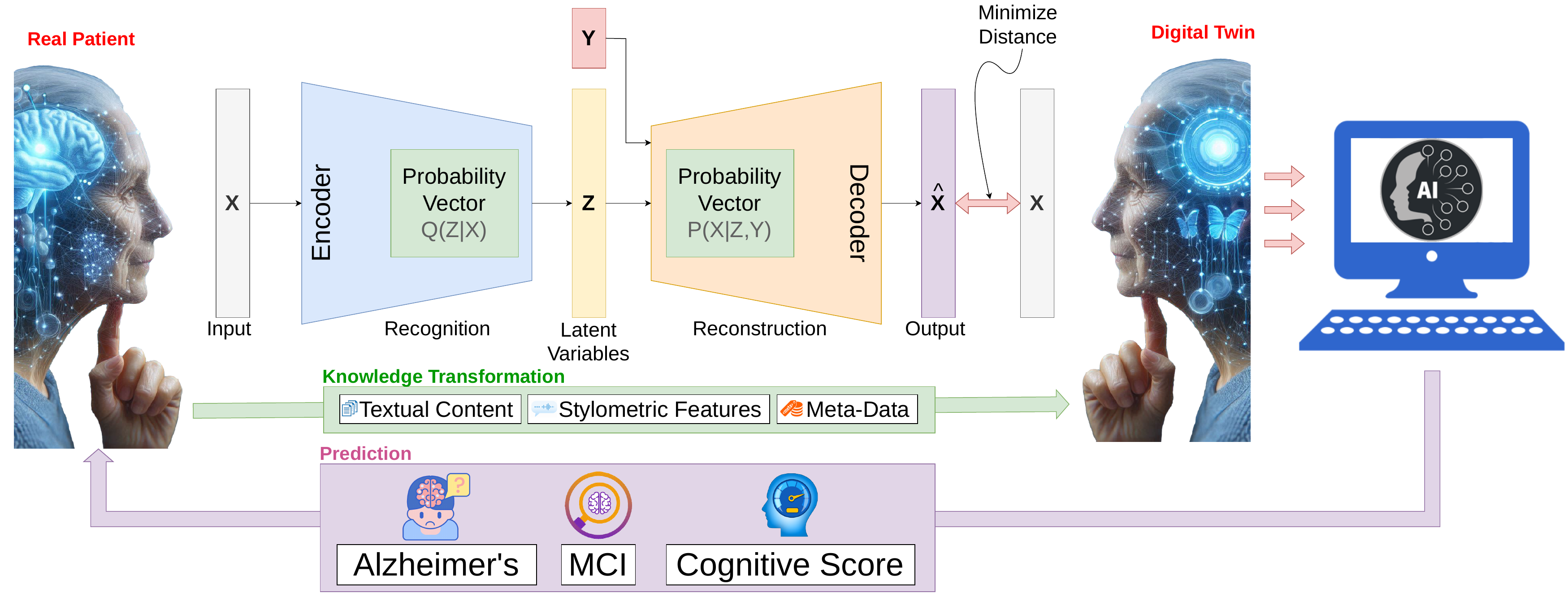}
\caption{Overview of the proposed language-based digital twin framework. The model integrates textual content, stylometric features (pause and tempo), and participant metadata to learn personalized conversational behavior and enable cognitive assessment.}
\label{proposal}
\end{figure}

To address this gap, we propose a \textit{language-based digital twin} framework that models individual conversational behavior using LLMs. The proposed approach captures linguistic style, temporal dynamics, and cognitive signatures by incorporating stylometric cues such as pause and tempo, along with contextual metadata, enabling the generation of personalized responses that resemble real human behavior. Fig.~\ref{proposal} illustrates the overall architecture of the proposed framework.

To evaluate response fidelity and cognitive consistency, we introduce a conditional variational autoencoder (cVAE)-based evaluator that measures reconstruction quality and predicts cognitive scores. Experiments on the I-CONECT dataset~\cite{dodge2024internet} demonstrate that the proposed digital twin preserves identity-specific characteristics and retains cognitively relevant information, enabling accurate estimation of cognitive status.

Overall, this work advances digital twin modeling from population-level representations toward individualized, language-centered approaches, enabling scalable and personalized monitoring of cognitive health in elderly populations. The implementation details are available at \href{https://github.com/MMHosseini}{GitHub}.
\section{Related Works}
Digital twins have emerged as a promising paradigm in healthcare, enabling personalized and dynamic computational representations of patients for monitoring, prediction, and decision support~\cite{tudor2025scoping,khoshfekr2025digital,nadeem2025comprehensive}. Unlike static predictive models, digital twins of individuals require continuous updating, personalization, and integration of heterogeneous data sources such as clinical records, behavioral signals, and sensor data~\cite{tudor2025scoping,nadeem2025comprehensive}. Recent advances in large language models (LLMs) further enable effective processing of unstructured and multimodal data, providing a strong foundation for adaptive digital twin systems.

In cognitive health, digital twins are particularly relevant due to the gradual and multifactorial nature of neurocognitive decline, where conditions such as Mild Cognitive Impairment (MCI) manifest across language, behavior, and daily activities. Prior work has explored digital twins for modeling cognitive states and predicting decline. For instance, Sprint et al.~\cite{sprint2024building} proposed HDTwin, integrating diverse cognitive-health signals using LLMs, while Lammert et al.~\cite{lammert2025large} developed an LLM-based framework for personalized clinical decision-making. Fard et al.~\cite{fard2024linguistic} introduced a Transformer-based approach to capture temporal linguistic patterns. Beyond cognitive health, digital twin systems have also been applied in domains such as cardiac modeling and precision medicine~\cite{corral2020digital,bjornsson2019digital}. 


A related line of work explores conversational agents and digital twins for cognitive support. A-CONECT~\cite{hong2024conect} presents an LLM-based system for dementia intervention, where personalized digital twins are trained on participant-specific conversational data to simulate patient interactions and evaluate conversational strategies at scale. However, the evaluation primarily focuses on overall conversational similarity, with limited investigation of long-term behavioral evolution and future cognitive trajectory modeling.

Despite these advances, existing digital twin frameworks largely focus on physiological modeling, prediction, or population-level simulation, with limited attention to fine-grained linguistic and behavioral patterns. In particular, prior works do not explicitly model conversational dynamics, stylistic features, or temporal interaction patterns at the individual level. 

In contrast, our approach introduces a \textit{personalized language-based digital twin} that models and reproduces individual conversational behavior, capturing linguistic style, temporal dynamics, and cognitive signatures. This shift from population-level or predictive modeling toward individualized behavioral emulation distinguishes our framework and enables more accurate and personalized modeling of cognitive states.
\section{Methodology}
\paragraph{\textbf{Digital Twin Formulation}:}
In this work, we conceptualize the language-based digital twin as a personalized, data-driven model that captures and reproduces an individual’s conversational behavior over time. Unlike traditional predictive models that map inputs to outputs, the digital twin aims to emulate underlying behavioral and cognitive patterns by leveraging longitudinal conversational data and metadata. This enables the model to generate realistic responses while reflecting individual-specific linguistic style, temporal dynamics, and cognitive characteristics. By continuously conditioning on participant-specific information, the proposed framework aligns with the core principles of digital twins, namely personalization, dynamic representation, and behavioral fidelity.

\subsection{Language Digital Twin}
\label{language_digital_twin}
We model a \textit{language-based digital twin} using an LLM to mimic the conversational style and linguistic behavior of an individual. We adopt \texttt{GPT-4.1-mini} as the base model and adapt it through supervised fine-tuning.

\paragraph{\textbf{Data Preprocessing and Stylometric Augmentation}:}
To capture individual speaking patterns, we augment transcripts with stylometric annotations reflecting temporal speech dynamics. Specifically, we introduce:
\begin{itemize}
    \item \textbf{PAUSE}: $\{\texttt{NONE}, \texttt{SHORT}, \texttt{MED}, \texttt{LONG}, \texttt{VLONG}\}$
    \item \textbf{TEMPO}: $\{\texttt{SLOW}, \texttt{MED}, \texttt{FAST}, \texttt{VFAST}\}$
\end{itemize}
These tokens encode timing and rhythm, allowing the model to learn both semantic content and temporal characteristics correlated with cognitive states~\cite{martinez2021ten}.

\paragraph{\textbf{Supervised Fine-Tuning (SFT)}:}
Training data is formatted as \textit{system}, \textit{user}, and \textit{assistant} messages. The system prompt defines the mimicry task, the user prompt includes the question and metadata (e.g., participant ID, age, gender, interview date, topic), and the assistant response corresponds to the participant’s answer enriched with stylometric tokens.

Given input $x = (q, m)$, where $q$ is the question and $m$ is metadata, the model is trained to generate response $y$:
\begin{equation}
\mathcal{L}_{\text{SFT}} = - \mathbb{E}_{(x,y)} \left[ \log p_{\theta}(y|x) \right].
\end{equation}

\subsection{cVAE-Based Evaluator}

We introduce a multi-head Conditional Variational Autoencoder (cVAE)~\cite{sohn2015learning} to evaluate response quality and cognitive alignment. The model assesses similarity between generated and real responses while predicting Mild Cognitive Impairment (MCI) scores.

\paragraph{\textbf{Model Architecture}:}
The cVAE is conditioned on question $q$, metadata $m$, and response $y$. The encoder maps these inputs to a latent representation:
\begin{equation}
q_{\phi}(z|q,m,y).
\end{equation}
The latent space is parameterized by mean $\mu$ and log-variance $\log \sigma^2$. The decoder reconstructs the response:
\begin{equation}
p_{\theta}(y|q,m,z).
\end{equation}

\paragraph{\textbf{Multi-Head Design}:}
A second head predicts an MCI-related score from $z$:
\begin{equation}
\hat{s} = f(z),
\end{equation}
where $f(\cdot)$ is a regression function.

\paragraph{\textbf{Loss Function}:}
Training combines reconstruction, KL divergence, and MCI prediction losses:
\begin{equation}
\mathcal{L}_{\text{rec}} = \mathbb{E}\left[\|y - \hat{y}\|^2\right],
\end{equation}
\begin{equation}
\mathcal{L}_{\text{KL}} = -\frac{1}{2} \mathbb{E}\left[1 + \log \sigma^2 - \mu^2 - \exp(\log \sigma^2)\right],
\end{equation}
\begin{equation}
\mathcal{L}_{\text{MCI}} = \mathbb{E}\left[\|s - \hat{s}\|^2\right],
\end{equation}
\begin{equation}
\mathcal{L} = \mathcal{L}_{\text{rec}} + \mathcal{L}_{\text{KL}} + \lambda \mathcal{L}_{\text{MCI}},
\end{equation}
where $\lambda$ controls the importance of cognitive prediction, with higher weight assigned to the MCI loss.

The cVAE acts as a \textit{challenger} by measuring how closely generated responses match real linguistic and cognitive patterns. Unlike standard similarity metrics, it provides joint evaluation of stylistic fidelity and cognitive relevance.
\section{Experimental Results}

\subsection{Dataset}
We utilize the I-CONECT dataset~\cite{dodge2024internet}, derived from a randomized clinical trial on conversational engagement in older adults (75+), including both cognitively normal and MCI participants. Unlike datasets based on constrained tasks~\cite{becker1994natural,luz2021alzheimer,luz2104detecting}, I-CONECT captures naturalistic, longitudinal conversations, making it suitable for modeling individual language patterns. It includes multimodal data such as transcripts, interaction dynamics, and metadata.

From approximately 70 participants, we select five individuals with the most sessions to ensure sufficient longitudinal data. This subset includes two males and three females aged 77–83, with MoCA scores ranging from 19 to 29. Repeated evaluations (every six months) enable modeling of temporal cognitive changes. MoCA (Montreal Cognitive Assessment) is a 30-point cognitive screening tool assessing domains such as memory, attention, language, and executive function, where lower scores indicate greater impairment.

\subsection{Data Preprocessing and Fine-tuning}
\begin{figure}[t]
\centering
\includegraphics[width=1.0\textwidth]{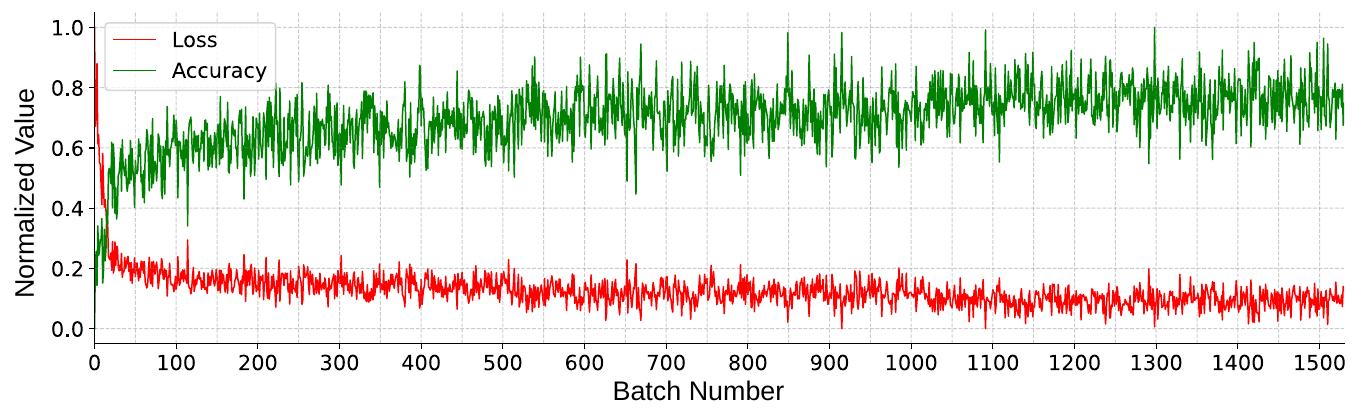}
\caption{Normalized loss and accuracy of the digital twin.} 
\label{sft_loss_and_accuracy}
\end{figure}

Original transcripts contained ASR errors; we reprocessed audio using Whisper~\cite{radford2023robust}. Speaker roles were separated using pyannote diarization~\cite{bredin2020pyannote}, ensuring accurate extraction of participant responses. Session-level embeddings were generated using Sentence-BERT~\cite{reimers2019sentence} and reduced via PCA to obtain topic descriptors for cVAE inputs. We fine-tune \texttt{GPT-4.1-mini} as the base model. Training shows stable convergence, with decreasing loss and increasing accuracy, as illustrated in Fig.~\ref{sft_loss_and_accuracy}.

\subsection{Feature Extraction}
We extract two feature types: embeddings and sentiment.

\paragraph{\textbf{Embedding Features}:}
Sentence-level embeddings are computed using the model \texttt{all-mpnet-base-v2}~\cite{reimers2019sentence}. Since each question or answer may contain multiple sentences, we aggregate sentence embeddings using mean and standard deviation to obtain fixed-length representations.

\paragraph{\textbf{Sentiment Features}:}
We use the \texttt{distilbert}~\cite{sanh2019distilbert} model to extract sentiment scores per sentence, aggregated using mean and standard deviation for both questions and answers.

\subsection{Feature Analysis}
We evaluate identity preservation using an SVM classifier to determine whether generated responses can be attributed to the correct participant.

Table~\ref{tbl:identity_detection} shows results across embedding, sentiment, and combined features for real, GPT, and digital twin responses. The digital twin achieves accuracy close to real data across all settings, while the base GPT model performs significantly worse (e.g., 44.15 vs. 48.55 vs. 19.90 for embedding Mean+STD). Similar trends are observed for other features.

These results indicate that the digital twin effectively captures individual linguistic patterns. Combining embeddings and sentiment yields the highest performance, particularly when using both mean and standard deviation.

\input{tables/identity_detection.tex}

\subsection{Digital Twin Evaluation}
To further evaluate the proposed language digital twin, we conduct an experiment focusing on response quality and cognitive (MoCA score) prediction. The cVAE model is trained on real question–answer pairs with participant metadata, including age, interview time, and contextual information. It is then used to evaluate responses from three sources: (1) real participant responses, (2) a base (raw) GPT model, and (3) the fine-tuned (FT) model representing the digital twin.

\begin{figure}[b]
\centering
\includegraphics[width=0.9\textwidth]{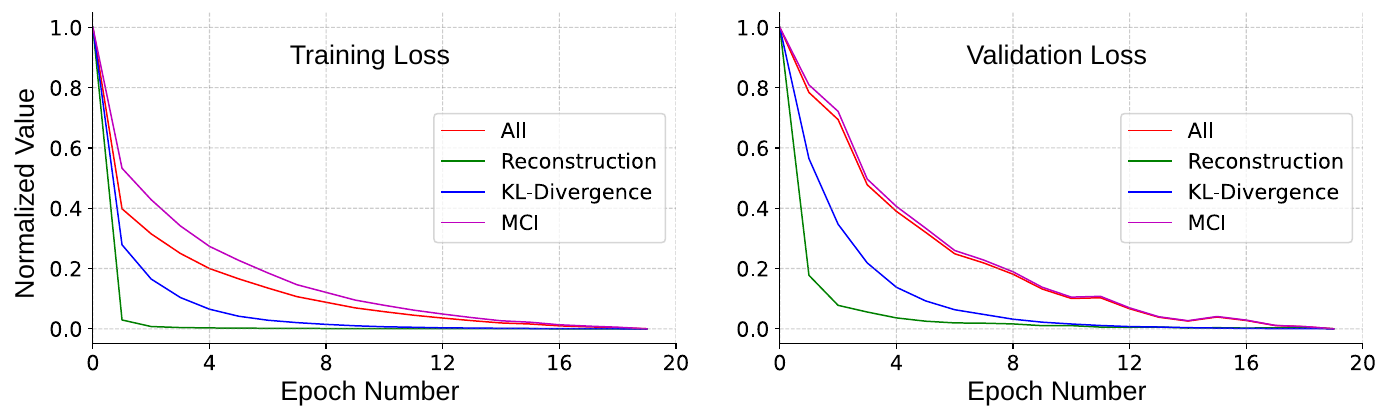}
\caption{Normalized loss function of the cVAE model for the first 20 epochs.} 
\label{cvae_loss}
\end{figure}

The cVAE serves a dual purpose: measuring reconstruction quality via Mean Squared Error (MSE) and predicting MoCA scores from the input question, generated response, and metadata, enabling joint evaluation of linguistic fidelity and cognitive consistency. Fig.~\ref{cvae_loss} illustrates the training dynamics of the cVAE model, showing a steady decrease in loss over the first 20 epochs, which indicates stable convergence.

\input{tables/reconstruction_error}

Table~\ref{tbl:reconstruction_error} reports reconstruction errors for real responses and the responses generated by digital twin across five participants. The digital twin closely matches real responses, with only slight increases (e.g., 0.0094 vs. 0.0098 for Participant~1), and errors remaining within a narrow range (0.0077–0.0098), indicating strong similarity and effective capture of linguistic patterns.

\input{tables/mci_error}

Table~\ref{tbl:mci_error} presents MoCA prediction errors for real responses, raw GPT, and digital twin outputs. The digital twin closely aligns with real data across participants (e.g., 0.94 vs. 0.92), while the base GPT shows substantially higher errors (3.53–5.08), indicating poor cognitive alignment.

Overall, the digital twin maintains low prediction errors (0.40–1.08), demonstrating its ability to preserve cognitively relevant information and captures linguistically and cognitally relevant patterns. These results highlight its potential as a non-invasive tool for continuous cognitive monitoring and early detection of cognitive decline.
\section{Conclusion and Future Works}
\paragraph{\textbf{Conclusion}:}
We introduced a language-based digital twin framework for modeling cognitive behavior in elderly individuals. By leveraging large language models and incorporating stylometric cues such as pause and tempo, the approach captures both linguistic style and temporal dynamics. Through supervised fine-tuning, the digital twin generates personalized responses that closely resemble real conversations.

To evaluate response fidelity and cognitive consistency, we proposed a cVAE-based framework that jointly measures reconstruction quality and cognitive prediction. Results show that the digital twin captures individual-specific linguistic patterns and achieves reconstruction and MoCA prediction errors comparable to real data, while outperforming baseline GPT-generated responses. These findings demonstrate that language-based digital twins can serve as a reliable and non-invasive tool for monitoring cognitive health.

\paragraph{\textbf{Future Work}:}
We plan to extend this framework to a multimodal digital twin by incorporating additional modalities from the I-CONECT dataset, such as audio and video. Integrating vocal features and facial expressions will enable improved modeling of affective and behavioral signals, leading to more accurate and comprehensive digital twins. This extension has the potential to enhance cognitive assessment and support robust monitoring of cognitive decline in real-world settings. A limitation of this study is the relatively small sample size; future work will involve evaluating the proposed framework on a larger and more diverse cohort to improve generalizability and robustness. Additionally, we will investigate generalization across participants to further assess the robustness of the proposed framework.

%
%
\section*{Acknowledgments}
This work was partially supported by a PennAITech award from the National Institute on Aging (Grant No. P30AG073105), the Knoebel Institute for Healthy Aging (KIHA), and the National Institute on Aging under Grants R01AG051628 and R01AG056102.
%
%
%
\bibliographystyle{splncs04}
\bibliography{ref}
\end{document}

%% file: tables/identity_detection.tex
\begin{table}[t!]
\centering
\caption{Identity detection accuracy (in \%) under different feature configurations. We consider three representations: \textit{Mean}, \textit{STD}, and \textit{All}. For each response, sentence-level features are first extracted and then aggregated using their mean (\textit{Mean}) and standard deviation (\textit{STD}) to form the response-level representation. In addition, the entire response is processed directly to obtain a single feature vector (\textit{All}). \textit{Mean}+\textit{STD} shows the joint feature vectors of \textit{Mean} and \textit{STD}.}
\label{tbl:identity_detection}
\setlength{\tabcolsep}{6pt}
\begin{tabular}{c|cccc}
\hline
 &  & Participant & GPT & Digital Twin \\ \hline

\multirow{4}{*}{Embedding} 
& Mean     & 48.35 & 19.44 & 42.90 \\
& STD      & 35.64 & 21.70 & 32.99 \\
& All      & 47.12 & 21.02 & 36.86 \\
& Mean+STD & 48.55 & 19.90 & 44.15 \\ \hline

\multirow{4}{*}{Sentiment} 
& Mean     & 50.64 & 24.31 & 41.17 \\
& STD      & 44.62 & 23.36 & 41.27 \\
& All      & 46.61 & 22.36 & 29.87 \\
& Mean+STD & 48.37 & 23.28 & 41.73 \\ \hline

\multirow{4}{*}{Embedding+Sentiment} 
& Mean     & 49.96 & 20.87 & 43.55 \\
& STD      & 42.83 & 22.24 & 37.56 \\
& All      & 49.27 & 22.25 & 37.54 \\
& Mean+STD & \textbf{50.95} & 21.51 & \textbf{44.42} \\ \hline

\end{tabular}
\end{table}

%% file: tables/reconstruction_error.tex
\begin{table}[t!]
\centering
\caption{Reconstruction error (ranging from 0 to 1) for real responses and responses generated by the digital twin. Low error values across all participants indicate that the digital twin closely reproduces the linguistic and contextual characteristics of real responses.}
\label{tbl:reconstruction_error}
\setlength{\tabcolsep}{8pt}
\begin{tabular}{l|ccccc}
\hline
Participant \#    & P1    & P2    & P3    & P4    & P5    \\ \hline
Real Response     & 0.0094 & 0.0078 & 0.0077 & 0.0078 & 0.0086 \\
Digital Twin      & 0.0098 & 0.0084 & 0.0094 & 0.0094 & 0.0089 \\

\hline
\end{tabular}
\end{table}

%% file: tables/mci_error.tex
\begin{table}[b!]
\centering
\caption{MoCA score prediction error for the real participants, their corresponding data generated by raw GPT and the digital twins.}
\label{tbl:mci_error}
\setlength{\tabcolsep}{10pt}
\begin{tabular}{l|ccccc}
\hline
Participant \#    & P1    & P2    & P3    & P4    & P5    \\ \hline
Real Participant & 0.94 & 0.58 & 1.05 & 0.40 & 1.03 \\
Raw GPT         & 3.53 & 5.08 & 4.60 & 4.95 & 4.59 \\
Digital Twin   & 0.92 & 0.55 & 1.06 & 0.41 & 1.08 \\
\hline
\end{tabular}
\end{table}